\documentclass[10pt,twocolumn,letterpaper]{article}

\usepackage{iccv}
\usepackage{times}
\usepackage{epsfig}
\usepackage{graphicx}
\usepackage{amsmath}
\usepackage{amssymb}

\usepackage{float}  
\usepackage{booktabs}
\usepackage{caption}
\usepackage{multirow} 
\usepackage{xcolor}
\usepackage{url} 
\usepackage{cite}
\usepackage{subcaption}
\usepackage{placeins}
\usepackage[group-separator={,}]{siunitx}
\usepackage[utf8]{inputenc}
\usepackage[export]{adjustbox}

\usepackage{siunitx}

\usepackage[pagebackref=true,breaklinks=true,letterpaper=true,colorlinks,bookmarks=false]{hyperref}

\newcommand{\bc}{\mathbf{c}}
\newcommand{\bd}{\mathbf{d}}

\newcommand{\bI}{\mathbf{I}}

\newcommand{\bK}{\mathbf{K}}

\newcommand{\bo}{\mathbf{o}}

\newcommand{\br}{\mathbf{r}}\newcommand{\bR}{\mathbf{R}}

\newcommand{\bt}{\mathbf{t}}

\newcommand{\bx}{\mathbf{x}}

\newcommand{\bz}{\mathbf{z}}

\newcommand{\bxi}{\boldsymbol{\xi}}

\newcommand{\nE}{\mathbb{E}}

\newcommand{\nR}{\mathbb{R}}
\newcommand{\nS}{\mathbb{S}}

\newcommand{\cD}{\mathcal{D}}

\newcommand{\cM}{\mathcal{M}}
\newcommand{\cN}{\mathcal{N}}

\newcommand{\cV}{\mathcal{V}}

\newcommand{\figref}[1]{Fig.~\ref{#1}}
\newcommand{\secref}[1]{Section~\ref{#1}}

\newcommand{\tabref}[1]{Tab.~\ref{#1}}

\makeatletter
\DeclareRobustCommand\onedot{\futurelet\@let@token\@onedot}
\def\@onedot{\ifx\@let@token.\else.\null\fi\xspace}
\def\eg{e.g\onedot} 
\def\ie{i.e\onedot}

\def\etal{et~al\onedot}

\makeatother

\newcommand{\boldparagraph}[1]{\vspace{0.2cm}\noindent{\bf #1:}}

\definecolor{darkgreen}{rgb}{0,0.7,0}

\iccvfinalcopy %
 
\def\iccvPaperID{*****} %

\ificcvfinal\fi

\renewcommand{\boldparagraph}[1]{\vspace{.1cm}\noindent\textbf{#1:}}
\newcommand{\boldquestion}[1]{\vspace{.1cm}\noindent\textbf{#1?}}
\newcommand{\boldparagraphnovspace}[1]{\vspace{-0.00cm}\noindent\textbf{#1:}}
\newcommand{\boldquestionnovspace}[1]{\vspace{-0.00cm}\noindent\textbf{#1?}}

\newcommand{\nRr}{\mathbb{R}^3}

\begin{document}

\title{CAMPARI: Camera-Aware Decomposed Generative Neural Radiance Fields}
\author{Michael Niemeyer$^{1,2}$ \quad Andreas Geiger$^{1,2}$\\
$^1$Max Planck Institute for Intelligent Systems, Tübingen \qquad $^2$University of Tübingen\\
{\tt\small \{firstname.lastname\}@tue.mpg.de}
}

\maketitle
\ificcvfinal\thispagestyle{empty}\fi

\begin{abstract}
    Tremendous progress in deep generative models has led to photorealistic image synthesis. While achieving compelling results, most approaches operate in the two-dimensional image domain, ignoring the three-dimensional nature of our world. Several recent works therefore propose generative models which are 3D-aware, \ie, scenes are modeled in 3D and then rendered differentiably to the image plane. This leads to impressive 3D~consistency, but incorporating such a bias comes at a price: the camera needs to be modeled as well. Current approaches assume fixed intrinsics and a predefined prior over camera pose ranges. As a result, parameter tuning is typically required for real-world data, and results degrade if the data distribution is not matched. Our key hypothesis is that learning a camera generator jointly with the image generator leads to a more principled approach to 3D-aware image synthesis. Further, we propose to decompose the scene into a background and foreground model, leading to more efficient and disentangled scene representations. While training from raw, unposed image collections, we learn a 3D- and camera-aware generative model which faithfully recovers not only the image but also the camera data distribution. At test time, our model generates images with explicit control over the camera as well as the shape and appearance of the scene.
 \end{abstract}

\section{Introduction}
\begin{figure}
    \centering  
    \begin{subfigure}[h]{\linewidth}
        \includegraphics[width=\linewidth]{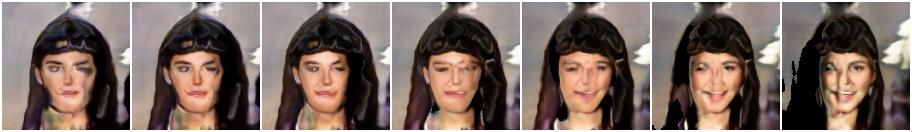}
        \includegraphics[width=\linewidth]{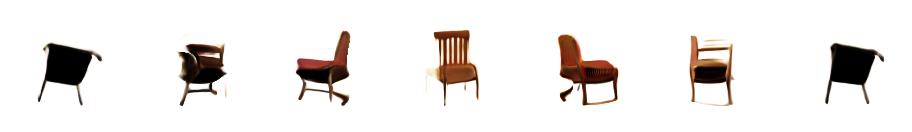}
        \includegraphics[width=\linewidth]{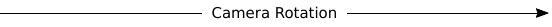}
        \vspace{-.6cm}
        \subcaption{3D-Aware Image Synthesis without Camera Tuning}
        \label{subfig:teasera}        
    \end{subfigure}
    \begin{subfigure}[h]{\linewidth} 
        \includegraphics[width=\linewidth]{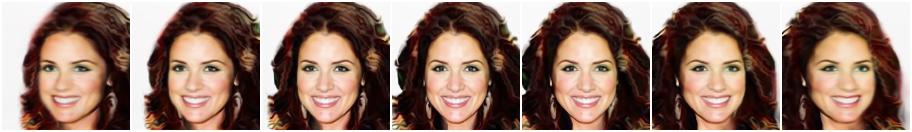}
        \includegraphics[width=\linewidth]{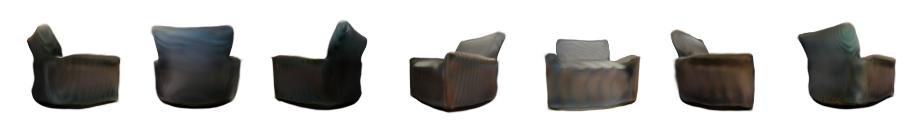}
        \includegraphics[width=\linewidth]{gfx/teaser_renderings/arrow}
        \vspace{-.6cm}
        \subcaption{3D- and Camera-Aware Image Synthesis (Ours)}
        \label{subfig:teaserb}        
    \end{subfigure}
    \begin{subfigure}[h]{\linewidth} 
        \includegraphics[width=\linewidth, trim={0 10 0 0}, clip]{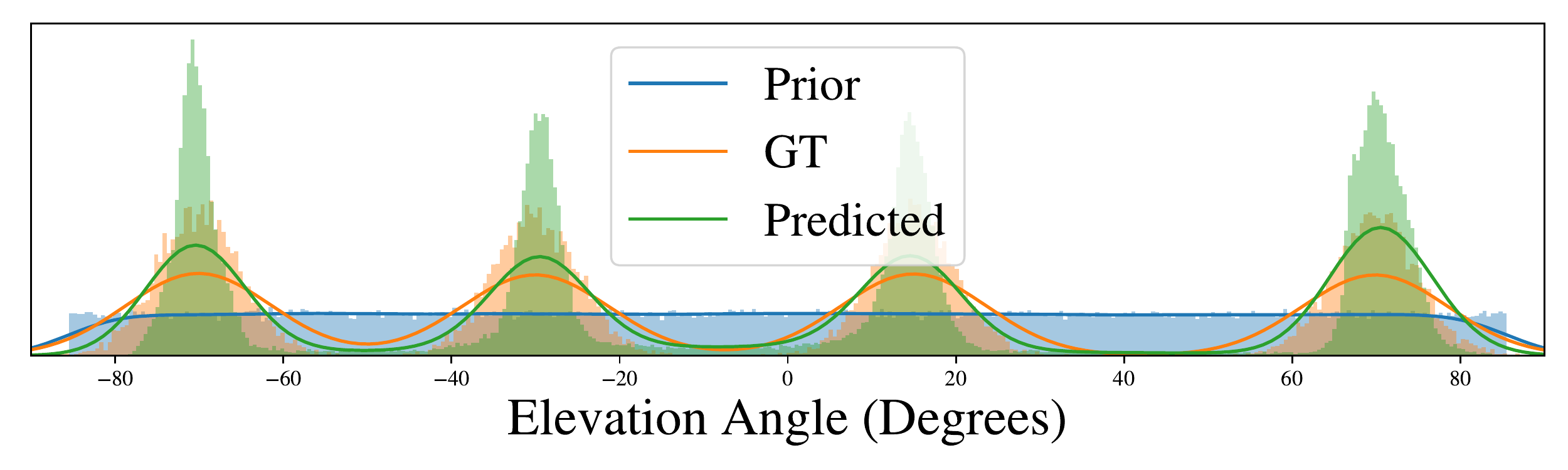}
        \vspace{-.5cm}
        \subcaption{Learned Camera Elevation Distribution for Chairs}
        \label{subfig:teaserc}        
    \end{subfigure}
    \vspace{-.3cm}
    \caption{
        \textbf{Overview.} 
        We propose to learn a 3D \textit{and camera}-aware generative model for controllable image synthesis.
        While 3D-aware models require tuned camera parameters and their results degrade if the pose distribution is not matched (\ref{subfig:teasera}), we learn a camera generator jointly with the image generator.
        While training from raw, unstructured image collections, we faithfully recover the camera distribution (\ref{subfig:teaserc}) and are able to generate 3D consistent representations (\ref{subfig:teaserb}) with explicit control over the camera viewpoint as well as the shape and appearance of the scene.
    }
    \label{fig:teaser}
\end{figure}
Deep generative models~\cite{Goodfellow2014NIPS,Kingma2014ICLR} are able to synthesize photorealistic images at high resolutions.
While state-of-the-art models~\cite{Karras2019CVPR, Karras2020CVPRi, Choi2018CVPR, Choi2020CVPR, Brock2019ICLR} produce impressive results, most approaches lack control over the generation process.
Control, however, is one key aspect required in many applications where generative models can be used.

To tackle this problem, many works investigate how architectures and training regimes can be improved to achieve more controllable image synthesis~\cite{Chen2016NIPS, Lee2020ARXIV, Peebles2020ECCV, Reed2014ICML, Karras2019CVPR, Zhao2018ECCV, Kwak2016ARXIV, Liu2020ARXIV, Goyal2019ARXIV, Zhu2015ICCV}.
Most approaches, however, operate on the two-dimensional image plane and hence do not consider the three-dimensional nature of our world.
While latent factors of variation representing \eg object rotation or translation may be found, full 3D disentanglement is hard to achieve.

In contrast, recent works~\cite{Schwarz2020NEURIPS,Henzler2019ICCV,Nguyen-Phuoc2019ICCVa,Chan2020ARXIV,Niemeyer2021CVPR,Nguyen-Phuoc2020NEURIPSp,Liao2020CVPR} incorporate three-dimensionality as an inductive bias into the generative model.
This leads to impressive, \textit{3D-aware} image synthesis in which the camera viewpoint can explicitly be controlled at test time. %
However, the incorporated inductive bias comes at a price: the camera needs to be modeled as well.
In practice, most works use fixed intrinsics and the true camera pose distribution for synthetic data or a uniform distribution over predefined ranges for real-world image collections.
As a result, these methods are either limited to simple data or typically require parameter tuning for real-world datasets.
Further, as a principled approach for obtaining the pose distribution is missing, results degrade if the distribution is not matched (see~\figref{subfig:teasera}).

\noindent\textbf{Contribution:} We propose Camera-Aware Decomposed Generative Neural Radiance Fields~(\textit{CAMPARI}), a novel generative model for 3D- and camera-aware image synthesis which is trained from raw, unposed image collections.
Our key idea is to learn a camera generator jointly with the image generator.
This allows us to apply our method to datasets with more complex camera distributions and, in contrast to previous works, requires no tuning of camera pose ranges.
We further propose to decompose the scene into a foreground and a background model, leading to a more efficient and disentangled scene representation.
We find that our model is able to learn 3D consistent representations (\figref{subfig:teaserb}) and to faithfully recover the camera distribution (\figref{subfig:teaserc}).
At test time, we can generate new scenes in which we have explicit control over the camera viewpoint as well the shape and appearance of the scene while training from raw, unposed image collections only.
\section{Related Work}\label{sec:rel-work}

\boldparagraphnovspace{Generative Adversarial Networks}
State-of-the-art Generative Adversarial Networks (GANs)~\cite{Goodfellow2014NIPS} allow for photorealistic image generation at high resolutions~\cite{Karras2019CVPR, Karras2020CVPRi, Choi2018CVPR, Choi2020CVPR, Brock2019ICLR, Anokhin2020ARXIV}.
As many applications require control mechanisms during image synthesis, 
a variety of works investigate how factors of variation can be disentangled without explicit supervision, \eg, by modifying the training objective~\cite{Peebles2020ECCV, Chen2016NIPS} or network architecture~\cite{Karras2019CVPR, Karras2020CVPRi}, or 
discovering factors of variation in latent spaces of pre-trained generative models~
\cite{Harkonen2020ARXIV, Collins2020CVPR, Jahanian2020ICLR, Shen2020CVPR, Goetschalckx2019ICCV, Abdal2020ARXIV, Zhan2020ARXIV}.
While achieving impressive results, the aforementioned works build on 2D-based convolutional or coordinate-based networks and hence model the image process in the two-dimensional image domain.
In this work, we advocate exploiting the fact that we know that our world is three-dimensional by combining a 3D generator with differentiable volume rendering. %

\boldparagraph{Neural Scene Representations}
Using coordinate-based neural representations to represent 3D geometry gained popularity in learning-based 3D reconstruction~\cite{Mescheder2019CVPR, Park2019CVPR, Chen2019CVPR,Oechsle2019ICCV, Saito2019ICCV,Niemeyer2019ICCV,Genova2019ICCV,Chen2019ICCV, Michalkiewicz2019ICCV} and several works~\cite{Liu2019NEURIPS,Niemeyer2020CVPR,Sitzmann2019NIPS,Yariv2020NEURIPS, Liu2020CVPR} propose differentiable rendering techniques for them.
Mildenhall~\etal~\cite{Mildenhall2020ECCV} propose Neural Radiance Fields (NeRFs) in which they combine a coordinate-based neural model with volume rendering for novel view synthesis.
We use radiance fields as 3D representation in our generative model due to their expressiveness and suitability for gradient-based learning.
While discussed methods require camera poses as input, recent works~\cite{YenChen2020ARXIV, Su2021ARXIV, Wang2021ARXIV} propose to estimate them instead. 
However, all aforementioned approaches fit the network weights to a single scene based on multi-view images of that scene, and do not have generation capabilities.
Our model, in contrast, allows for controllable image synthesis of generated scenes and is trained from unstructured image collections with only a single, unposed image per scene.

\boldparagraph{3D-Aware Image Synthesis}
Many recent approaches investigate how a 3D representation can be incorporated into the generator model~\cite{Henzler2019ICCV, Lunz2020ARXIV, Nguyen-Phuoc2019ICCV, Nguyen-Phuoc2020NEURIPSp, Liao2020CVPR, Schwarz2020NEURIPS, Gadelha2017THREEDV, Rezende2016NIPS, Henderson2020ARXIV, Henderson2020CVPR, Henderson2019IJCV, Niemeyer2021CVPR, Chan2020ARXIV}.
While some use additional supervision~\cite{Zhu2018NIPS, Wang2016ECCV, Alhaija2018ACCV, Chen2020ARXIV, Wu2016NIPS}, in the following, we focus on methods that are trained from unposed image collections similar to our approach. 
While voxel-based representations~\cite{Henzler2019ICCV} in combination with differentiable rendering lead to 3D controllable image synthesis, the visual quality is limited due to voxels' cubic memory growth.
Voxelized feature grids~\cite{Nguyen-Phuoc2019ICCV,Nguyen-Phuoc2020NEURIPSp, Liao2020CVPR} with neural 2D rendering lead to impressive results, but 
training is less stable and results less multi-view consistent due to the learnable projection. %
Very recently, methods~\cite{Schwarz2020NEURIPS, Chan2020ARXIV,Niemeyer2021CVPR} have been proposed which use radiance fields as the underlying representation similar to our approach.
However, all aforementioned methods use fixed camera instrinsics and pose distributions over predefined rotation, elevation, and translation ranges.
While this allows for 3D consistent image synthesis on synthetic data, it requires tuning for real-world datasets and results degrade if the real data distribution is not matched (see~\figref{subfig:teasera}).
In contrast, we learn a 3D- and camera-aware generative model by jointly estimating 3D representations and camera distributions.

\section{Method}
\begin{figure*}
    \centering
    \includegraphics[width=\linewidth]{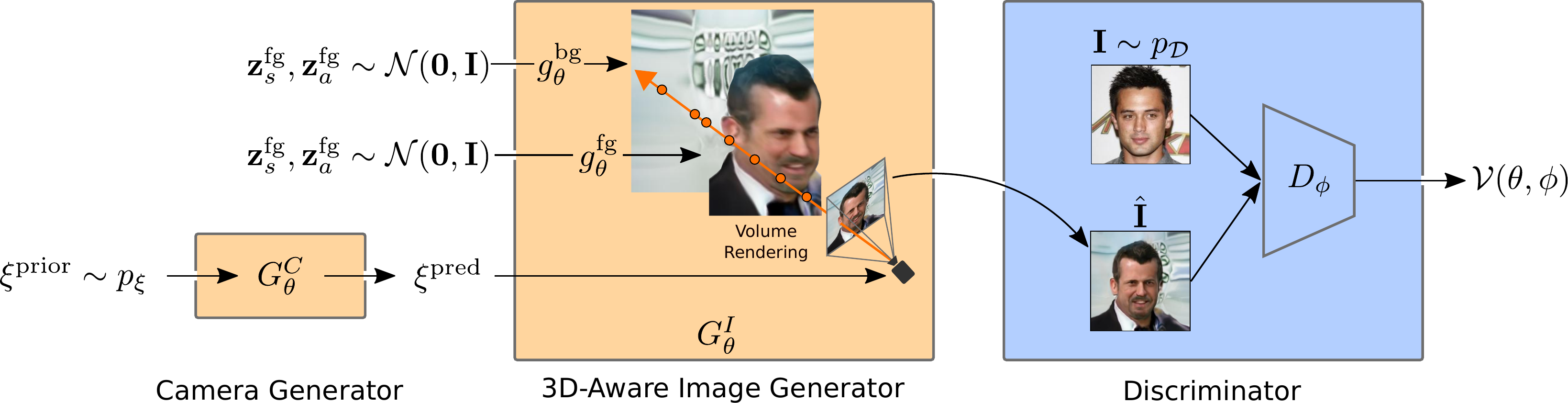}
    \caption{
        \textbf{CAMPARI.}
        During training, we sample a prior camera $\bxi^\text{prior} \sim p_\xi$ and pass it to our camera generator $G_\theta^C$ which predicts camera $\bxi^\text{pred}$. 
        Next, we pass the predicted camera together with latent shape and appearance codes $\bz^\text{fg}_s, \bz^\text{bg}_s, \bz^\text{fg}_a,  \bz^\text{bg}_a \sim \cN(\mathbf{0}, \mathbf{I})$
        to our 3D-aware image generator $G_\theta^I$.
        We then differentiably render the image $\hat{\mathbf{I}}$ of the scene for camera $\bxi^\text{pred}$ using volume rendering.
        Finally, our discriminator $D_\phi$ takes as input the generated image $\hat{\bI}$ and real image $\bI$ drawn from the data distribution $p_{\cD}$ and predicts whether they are real or fake.
        While training from raw image collections,
        at test time, we have explicit control over the camera and latent shape and appearance codes allowing for 3D- and camera-aware image synthesis.
    }
    \label{fig:method-overview}
\end{figure*}

Our goal is a 3D- and camera-aware generative model that is trained from raw image collections.
We first discuss our 3D-aware image generator model in~\secref{sec:graf}.
Next, we describe how a camera generator can be learned jointly to train a 3D- and camera-aware generative model in~\secref{sec:camera-pose}.
Finally, we describe our training procedure and implementation details in~\secref{sec:training} and~\ref{sec:impl-details}.
\figref{fig:method-overview} contains an overview of our method.

\subsection{3D-Aware Image Generator}\label{sec:graf}
\begin{figure}
    \centering
    \includegraphics[width=1.\linewidth]{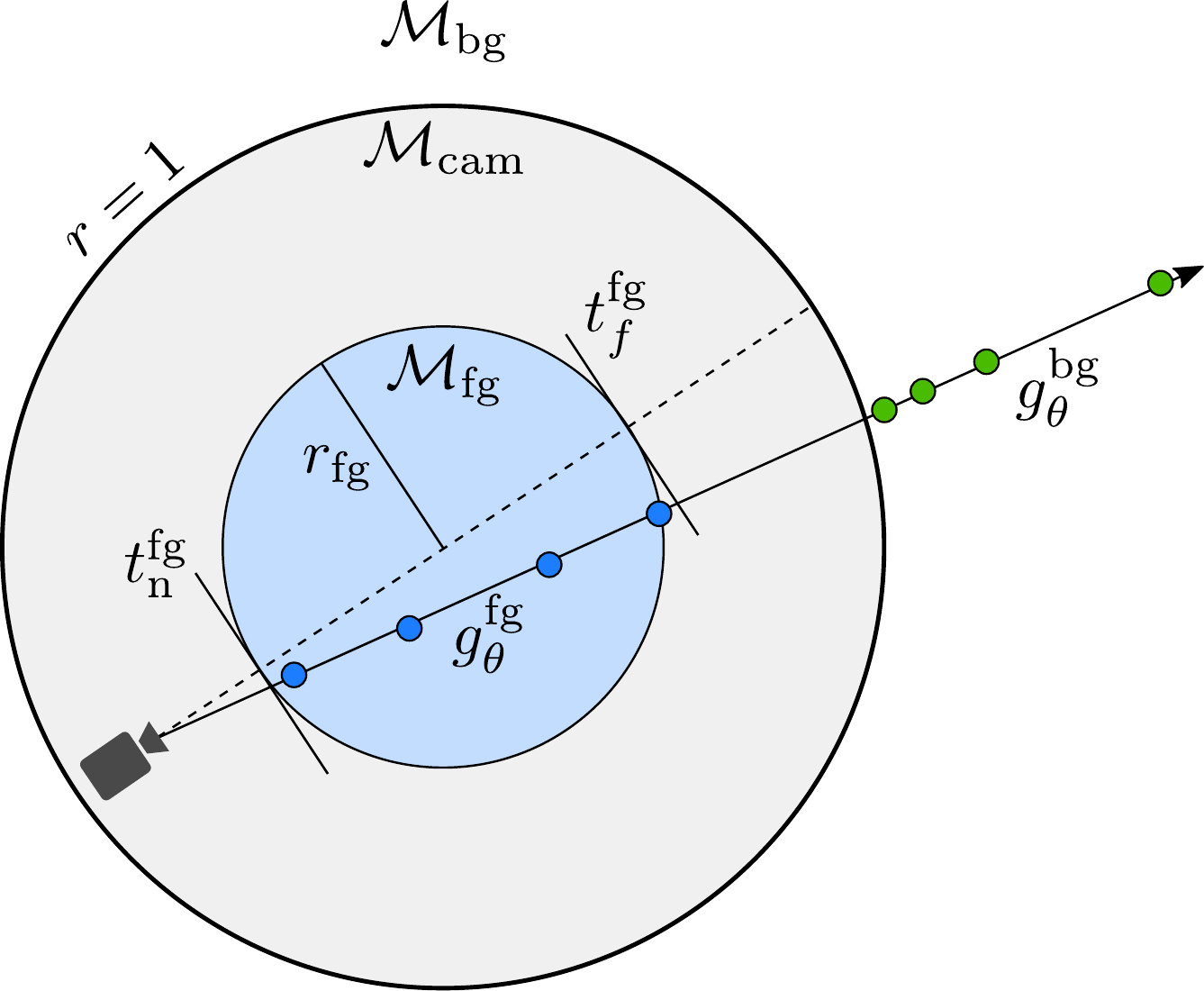}
    \caption{
        \textbf{Scene Space Sampling.}
        We assume the foreground $\cM_\text{fg}$ to be roughly inside a sphere of radius $r_\text{fg}$ and sample uniformly within resulting near and far bounds $(t_n^\text{fg}, t_f^\text{fg})$ for $g_{\theta}^\text{fg}$ (blue points).
        The background is assumed to be outside the unit sphere and we sample uniformly in \textit{inverse depth} for $g_{\theta}^\text{bg}$ (green points).
        }
    \label{fig:decomposed-scene}
\end{figure}

\boldparagraphnovspace{Scene Representation}
Our goal is to learn a generative model of single object scenes with background.
This allows us to incorporate prior knowledge into the scene representation for more efficient and disentangled representations. 
We represent scenes using a foreground and a background model.
More specifically, we partition 3D space $\nRr$ into a foreground $\cM_\text{fg} \subset \nRr$ and a background $\cM_\text{bg} \subset \nRr$, where the foreground is encapsulated by a sphere\footnote{In practice, we enforce the foreground to only be roughly inside the sphere of radius $r_\text{fg}$ as we sample within the same near and far bounds for all pixels of the same image (see~\figref{fig:decomposed-scene}).} of radius $r_\text{fg}<1$
\begin{align}
    \cM_\text{fg} = \left\{ \bx \in \nRr \mid {{\Vert\bx\Vert}_2} \leq r_\text{fg} \right\}
\end{align}
and the background is everything outside the unit sphere
\begin{align}
    \cM_\text{bg} =  \left\{ \bx \in \nRr \mid 1 \leq {{\Vert\bx\Vert}_2} \right\}
\end{align}
as illustrated in~\figref{fig:decomposed-scene}. 
We define the space of possible camera locations to be the space between fore- and background
\begin{align}
    \cM_\text{cam} =  \left\{ \bx \in \nRr \mid r_\text{fg} < {{\Vert\bx\Vert}_2} < 1 \right\}
\end{align}
As a result, the foreground is in front, and the background behind every possible camera.
In contrast to~\cite{Zhang2020ARXIV2} where the foreground is assumed to be within the unit sphere, our representation exhibits a stronger bias for single-object scenes while not being limited to specific scenarios as we do not enforce hard constraints.

\boldparagraph{Object Representation}
To represent the fore- and background, we use conditional neural radiance fields~\cite{Mildenhall2020ECCV, Schwarz2020NEURIPS} which are multilayer perceptrons (MLPs) mapping a 3D point $\bx \in \nRr$ and viewing direction $\bd \in \nS^2$ together with latent shape and appearance codes $\bz_s, \bz_a \in \nR^{L_z}$ to a density $\sigma \in \nR^+$ and RGB color $\bc \in \nRr$:
\begin{align}
    \begin{split}
        g_\theta: \nR^{L_\bx} \times \nR^{L_\bd} \times \nR^{L_z} \times \nR^{L_z} &\to \nR^+ \times \nRr \\
        (\gamma(\bx), \gamma(\bd), \bz_s, \bz_a) &\mapsto (\sigma, \bc)        
    \end{split}
\end{align}
where $\theta$ indicates the network parameters,  $\gamma$ the positional encoding~\cite{Mildenhall2020ECCV, Tancik2020NEURIPS} applied element-wise to $\bx$ and $\bd$, $L_x, L_d$ the output dimensions of the positional encodings, and $L_z$ the latent code dimension.
We use two separate networks for the fore- and background, and define our 3D scene representation as 
\begin{align}\label{eq:decomposed-net}
    \begin{split}
        g_\theta(\bx, \bd, \bz_s, \bz_a) = \begin{cases}
            g_{\theta}^\text{fg}(\bx, \bd, \bz_s, \bz_a), & \text{for } \bx \in \cM_\text{fg}\\
            g_{\theta}^\text{bg}(\bx, \bd, \bz_s, \bz_a), & \text{for } \bx \in \cM_\text{bg}
            \end{cases}
    \end{split}
\end{align}
To avoid cluttered notation, we always use the same $\theta$ to indicate network parameters. 

\boldparagraph{Scene Rendering}
To render images of our scene representation, classic volume rendering techniques can be used~\cite{Mildenhall2020ECCV, Kajiya1984SIGGRAPH} which are trivially differentiable. 
More specifically, for given camera intrinsics $\bK$ and extrinsics $[\bR \vert \bt]$, a pixel's color value $\bc_\text{final}$ is calculated by integrating over the camera ray $\br(t) = \bo + t\bd$ within near and far bounds $t_n, t_f$
\begin{align}\label{eq:int}
    \begin{split}
        \bc_\text{final} = \int_{t_n}^{t_f} T(t) \, \sigma(\br(t)) \, \bc(\br(t), \bd) \,dt \\
        \text{where} \quad T(t) = \exp \left( -\int_{t_n}^t \sigma(\br(s)) \,ds \right)
    \end{split}
\end{align}

\boldparagraph{Scene Spacing Sampling}
As no analytical solution exists for integral~\ref{eq:int}, it is commonly approximated using stratified sampling~\cite{Mildenhall2020ECCV}.
To faithfully render the scene, however, it is crucial that the numerical integration approximates the true integral well.
As we decompose the scene into fore- and background, we are able to inject prior knowledge into the sampling process, thereby saving computational cost and encouraging disentanglement.

More specifically, for camera pose $[\bR \vert \bt]$, we sample points uniformly for the foreground within bounds
\begin{align}
    (t_n^\text{fg}, t_f^{\text{fg}}) = \left( {\Vert\bt\Vert}_2 - r_\text{fg},  {\Vert\bt\Vert}_2 + r_\text{fg} \right)
\end{align}
For the background, we adopt the inverted sphere parametrization from~\cite{Zhang2020ARXIV2} where a 3D point $\bx$ outside the unit sphere is described as 
\begin{align}
    \bx^\prime = \left( \frac{\bx}{ {\Vert\bx\Vert}_2}, \frac{1}{{\Vert\bx\Vert}_2} \right) \in \left[ -1, 1 \right]^4
\end{align} 
We then sample points uniformly between $0$ and $1$ in \textit{inverse depth} for the background~\cite{Zhang2020ARXIV2}.
This way, we do not need to assume the background to be within a predefined range, but sample space denser if it is nearer to the foreground.

\boldparagraph{3D-Aware Image Generator}
We define our image generator $G^I_\theta$ as a function which renders an image of $g_\theta$ for given camera $\bxi = (\bK, [\bR\vert\bt])$ and fore-and background shape and appearance codes $\bz = \{ \bz^\text{fg}_s, \bz^\text{bg}_s, \bz^\text{fg}_a, \bz^\text{bg}_a\}$
\begin{align}
    {\hat \bI} = G^I_\theta(\bxi, \bz)
\end{align}
where $\hat\bI$ denotes the generated image.

\subsection{Camera Generator}\label{sec:camera-pose}
While incorporating a 3D representation into the generator leads to more controllable image synthesis, it also comes at a price: the camera and its pose distribution needs to be modeled as well.
A key limitation of state-of-the-art 3D-aware image synthesis models~\cite{Nguyen-Phuoc2019ICCVa, Henzler2019ICCV,Schwarz2020NEURIPS,Nguyen-Phuoc2020NEURIPSp, Niemeyer2021CVPR, Chan2020ARXIV} is that camera intrinsics and the pose distribution are predefined. %
This requires tuning of the range parameters and leads to degraded results if the camera distribution is wrong.
In the following, we describe how a camera generator can be learned jointly with the image generator to avoid tuning and to improve results if the camera distribution is unknown.

\boldparagraph{Camera Intrinsics}
Assuming a pinhole camera model, we can express the intrinsics as
\begin{align}
    \begin{split}
        \bK = \begin{pmatrix}
            f_x & 0   & c_x\\
            0   & f_y & c_y\\
            0   & 0   & 1
        \end{pmatrix}
    \end{split}
\end{align}
where $f_x, f_y$ indicate the focal lengths and $(c_x, c_y)^T$ the principal point.
In this work, we assume the principal point to lie in the center of the image plane, and hence $(c_x, c_y) = (\frac{W}{2}, \frac{H}{2})$ where $H \times W$ indicates the image  resolution.
As a result, the camera intrinsics are reduced to
\begin{align}
    \bxi^{\text{intr}} = (f_x, f_y) \in \nR^2
\end{align}

\boldparagraph{Camera Pose}
We parameterize the camera pose as a location on a sphere with radius $r_\text{cam}$ looking at the world-space origin $(0, 0, 0)^T \in \nRr$ and fixing the up-right position.
The camera pose can hence be easily described using radius $r_\text{cam}$, a rotation angle $\alpha_r \in \left[ -\pi, \pi \right]$ and an elevation angle $\alpha_e \in [-\frac{\pi}{2}, \frac{\pi}{2}]$:
\begin{align}
    \bxi^\text{pose} = (r_\text{cam}, \alpha_r, \alpha_e) \in \nRr
\end{align}
We obtain $\left[\bR\vert\bt\right]$ from $\bxi^\text{pose}$ as the composition 
of the Euler rotation matrices resulting from $\alpha_r$ and $\alpha_e$, respectively, and the translation vector resulting from $r_\text{cam}$.
When operating on $360^\circ$ rotation scenes, we represent $\alpha_r$ by a $2\times2$ matrix and project it to the special orthogonal group $SO(2)$ avoiding periodic boundary issues and ensuring well-behaved gradients~\cite{Levinson2020NEURIPS}.

\boldparagraph{Camera Generator}
\begin{figure}
    \centering
    \begin{subfigure}[h]{\linewidth}
        \includegraphics[width=\linewidth]{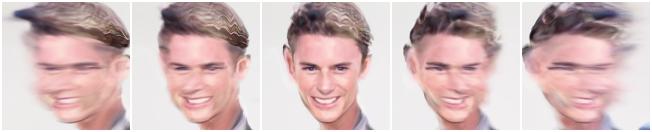}
        \caption{Camera Generator without Residual Design}\label{subfig:nores}
    \end{subfigure}
    \begin{subfigure}[h]{\linewidth}
        \includegraphics[width=\linewidth]{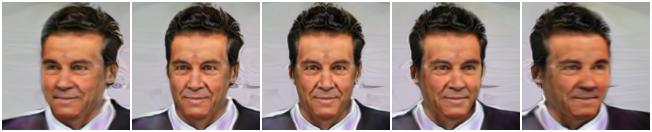}
        \caption{Camera Generator with Residual Design (Ours)}\label{subfig:res}
    \end{subfigure} 
    \vspace{-.08cm}
    \caption{
        \textbf{Residual Camera Generator.}
        We show camera rotation in a fixed range for our model using a camera generator with and without a residual connection to the sampled input camera.
        The residual design encourages exploration of a larger pose range from the beginning of training leading to more 3D consistent results.
    }
    \label{fig:residual_cam}
\end{figure}
Our goal is to learn a camera generator $G_\theta^C$ in addition to the image generator to obtain a 3D- and camera-aware generative model.
We define a camera as 
\begin{align}
    \bxi =  (\bxi^\text{intr}, \bxi^\text{pose}) %
\end{align}
We implement $G_\theta^C$ as an MLP which maps a \textit{prior} camera $\bxi^\text{prior} \sim p_{\xi}$ to a predicted camera $\bxi^\text{pred} \in \nRr$.
While the input to our camera generator could be of any dimension, using the same space as input and output allows us to incorporate prior knowledge and to design $G_\theta^C$ as a residual function~\cite{Ehrhardt2020ARXIV, He2016CVPR}.
More specifically, we define
\begin{align}\label{eq:gcam}
    G_\theta^C(\bxi^\text{prior}) = \bxi^\text{prior} + \Delta\bxi_\theta \quad\text{where}\quad \bxi_\text{prior} \sim p_\xi
\end{align}
and $\theta$ indicates the network parameters.
This way, we are able to encode prior knowledge into the prior distribution $p_\xi$ which encourages our model to explore a wider range of camera poses directly from the start of training.
It is important to note that as we train from raw, unposed image collections, our generator is not forced to learn 3D consistent representations and could therefore predict a static camera.
We find that the residual design is key to avoid trivial solutions like this (see~\figref{fig:residual_cam}).
In practice, we set $p_\xi$ to a Gaussian or uniform prior, however, more complex prior distributions can be incorporated as well.

\subsection{Training}\label{sec:training}

\boldparagraphnovspace{Progressive Growing}
To avoid excessive memory requirements and to improve training stability, we train our models using progressive growing~\cite{Karras2018ICLR,Chan2020ARXIV}.
We start training on a low image resolution which allows the generator and discriminator to focus on coarse structures during early iterations and to train with higher batch sizes.
As training progresses, we increase the image resolution until we reach the final resolution ($128^2$ pixels).
Due to GPU memory limitations, we reduce the batch size in each progressive growing step.

\boldparagraph{Discriminator} We implement the discriminator using residual blocks~\cite{He2016CVPR} of CoordConv layers~\cite{Liu2018NIPS} similar to~\cite{Chan2020ARXIV}.  
We progressively add new residual blocks when a new progressive growing step is reached.
We follow~\cite{Karras2018ICLR,Chan2020ARXIV} and fade in newly added layers to allow for a smooth transition.%

\boldparagraph{Training} During each iteration, we first sample a camera from the prior $\bxi_\text{prior} \sim p_\xi$.
Next, we pass the sampled prior camera to our camera generator $G_\theta^C$ and obtain the predicted camera $\bxi_\text{pred}$.
Finally, we volume render the predicted image from camera $\bxi_\text{pred}$ and for sampled latent shape and appearance codes $\bz = \{\bz^\text{fg}_s, \bz^\text{bg}_s, \bz^\text{fg}_a, \bz^\text{bg}_a\}$ which are all drawn from a unit Gaussian $\cN(\mathbf{0}, \mathbf{I})$.
We train our model with the non-saturating GAN objective~\cite{Goodfellow2014NIPS} and $R_1$ gradient penalty~\cite{Mescheder2018ICML}:
\begin{align}
    \begin{split}
        &\cV(\theta, \phi) = \\
        &\nE_{\bxi_\text{prior}\sim p_\xi, \bz \sim p_z} \left[ f(D_\phi(G^I_\theta(G_\theta^C(\bxi_\text{prior}), \bz))) \right] +\\
        &\nE_{\bI \sim p_\cD}\left[ f(-D_\phi(\bI)) - \lambda ||\Delta D_\phi(\bI)||^2 \right]
    \end{split}
\end{align}
where $f(t) = -\log(1 + \exp(-t))$, $\lambda = 10$, and $p_\cD$ represents the data distribution.

\subsection{Implementation Details}\label{sec:impl-details}
\boldparagraphnovspace{Network Parametrization}
We parameterize our radiance fields $g_{\theta}^\text{fg}$ and $g_{\theta}^\text{bg}$ as MLPs with $8$ hidden layers of dimension $128$, ReLU activation, and a skip connection to the fourth layer~\cite{Park2019CVPR}.
We concatenate the latent shape and appearance codes to the encoded input point and viewing direction, respectively~\cite{Schwarz2020NEURIPS}.
We implement our camera generator $G_\theta^C$ as an MLP with $4$ hidden layers of dimension $64$ and ReLU activation, and we initialize the last layer's biases as zeros and the weights from $\cN(0, 0.05)$.
After adding the learned offset to the prior pose~\eqref{eq:gcam}, we clamp the output to be within a valid range (see sup.\ mat.).
We parameterize our discriminator using $5$ residual blocks~\cite{He2016CVPR} of CoordConv layers~\cite{Liu2018NIPS} with leaky ReLU activation similar to~\cite{Chan2020ARXIV}.  

\boldparagraph{Training Procedure}
We schedule the number of sample points along the ray~\cite{Niemeyer2019ICCV} and, depending on the scene type, sample between $20$ and $52$ points on each ray at the final stage.
We use the RMSprop~optimizer~\cite{Tieleman2012Coursera} with learning rates of \num{5e-4} and \num{1e-4} for our generators and discriminator, respectively.
We use exponential learning rate decay with a rate of $0.1$ after \num{1.5e5} iterations~\cite{Mildenhall2020ECCV}.
For the generator weights, we use an exponential moving average~\cite{Yazici2019ICLR} with decay $0.999$.
To ensure training stability, we fix our camera generator for the first iterations. %
We start progressive growing at $32^2$ pixels and double the resolution after \num{2e4} and \num{7e4} iterations.
We use batch sizes of $[64, 24, 20]$ for $180^\circ$ and $[64, 20, 15]$ for $360^\circ$ rotation scenes. %
We train on a single NVIDIA V100 GPU.

\section{Experiments}

\boldparagraphnovspace{Datasets}
We run experiments on the commonly-used datasets \textit{Cats}~\cite{Zhang2003ICCV}, \textit{CelebA}~\cite{Liu2015ICCV}, and \textit{Cars}~\cite{Schwarz2020NEURIPS}.
In contrast to previous works on 3D-aware image synthesis~\cite{Nguyen-Phuoc2019ICCVa, Schwarz2020NEURIPS} we use a center crop of the entire image for CelebA instead of a close-up region.
Note that learning a consistent 3D representation becomes more challenging as the data variety is larger and ideally the model should disentangle fore- and background.
We further create synthetic datasets \textit{Chairs1} and \textit{Chairs2} which consist of photorealistic renderings of the Photoshape~chairs~\cite{Park2018ACM}.
To test our method on complex camera pose distributions, we sample rotation and elevation angles from mixtures of Gaussians (see sup.\ mat.\ for details).
For Cats and the synthetic datasets we only use a foreground model as they do not contain any background.

\boldparagraph{Baselines}
We compare against the state-of-the-art 3D-aware methods HoloGAN~\cite{Nguyen-Phuoc2019ICCV} and GRAF~\cite{Schwarz2020NEURIPS} which are both suited for single-object scenes like our approach.
In HoloGAN, scenes are represented as voxelized feature grids which are differentiably rendered via a reshaping operation and learnable convolutional filters.
Similar to us, GRAF uses radiance fields as 3D representation.
Note that while for both methods, the goal is 3D-aware image synthesis, we also decompose the scene into fore- and background. 
Further, in contrast to us, both methods require hand-tuned cameras.
We therefore report results for them in tuned and non-tuned settings where for the first, we use the ranges reported by the authors, and for the latter, we use the ranges of our prior distribution (see sup.\ mat.\ for details).

\boldparagraph{Camera Distributions}
For Cats and CelebA, we use a Gaussian prior for both rotation and elevation, and for Carla and Chairs, we use a uniform distribution over the entire rotation and elevation range.
For the camera radius and focal lengths, we use a Gaussian prior except for Chairs where we fix the focal length.
Note that jointly optimizing camera intrinsics and extrinsics has many different valid solutions, but as we are interested in comparing our results against the ground truth, we fix the intrinsics.

\boldparagraph{Metrics}
We adhere to common practice and report the Frechet Inception Distance (FID) 
to quantify image quality.

\subsection{Results}

\boldquestionnovspace{How does our approach compare to baseline methods}
\begin{table}
    \centering
    \resizebox{1.\linewidth}{!}{
    
\begin{tabular}{l|c|ccccc}
    \toprule
    & Tuned? & Cats & CelebA & Carla &Chairs1 & Chairs2\\ 
    \midrule
    \multirow{2}{*}{HGAN~\cite{Nguyen-Phuoc2019ICCV}} & yes & 34 & 67 & 153 & - & - \\
    & no & 42 & 83 & 169 & 124 & 116 \\
    \multirow{2}{*}{GRAF~\cite{Schwarz2020NEURIPS}} & yes & 21 & 38 & 28 & - & - \\
    & no & 72 & 74 & 90 & 55 & 53 \\
    Ours & no & 23 & 28 & 39 & 31 & 33 \\
    \bottomrule
\end{tabular}

    }
    \caption{
        \textbf{Quantitative Comparison.}
        We report FID ($\downarrow$) for baselines and our method at $128^2$ pixel resolution.
    }
    \label{tab:tab128}
\end{table}
\begin{figure}
    \begin{subfigure}[h]{\linewidth}
        \includegraphics[width=\linewidth]{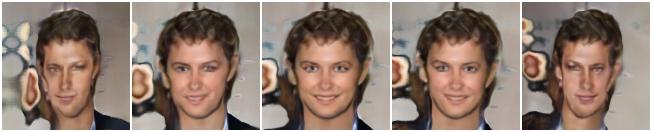}
        \caption{Rotation for GRAF~\cite{Schwarz2020NEURIPS} (Camera Parameters Tuned)}
        \label{subfig:compa}
    \end{subfigure}
    \begin{subfigure}[h]{\linewidth}
        \includegraphics[width=\linewidth]{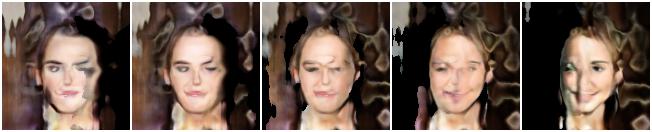}
        \caption{Rotation for GRAF~\cite{Schwarz2020NEURIPS} (Camera Parameters not Tuned)}
        \label{subfig:compb}
    \end{subfigure}
    \begin{subfigure}[h]{\linewidth}
        \includegraphics[width=\linewidth]{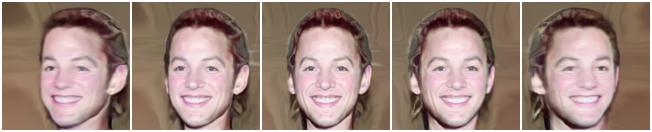}
        \caption{Rotation for Ours (No Tuning Required)}
        \label{subfig:compc}
    \end{subfigure}
    \caption{
        \textbf{Qualitative Comparison.}
        State-of-the-art 3D-aware models typically require hand-tuned camera parameters (\figref{subfig:compa}), and results degrade if the data distribution is not matched (\figref{subfig:compb}).
        In contrast, we learn a camera generator jointly with the image generator leading to more 3D consistent results (\figref{subfig:compc}) while no tuning is required. 
    }
    \label{fig:qual-comp}
\end{figure}
In~\tabref{tab:tab128} and \figref{fig:qual-comp} we show quantitative and qualitative comparisons of our method to baselines.
Although our model does not require tuning of camera parameters, we achieve similar or better performance on datasets for which baseline methods are tuned.
We find that the results of baselines indeed depend on tuned camera parameters, and their performances drop if the data's camera distribution is not matched.
We further observe that the performance drop is more significant for GRAF~\cite{Schwarz2020NEURIPS} than for HoloGAN~\cite{Nguyen-Phuoc2019ICCV} which we attribute to HoloGAN's learnable projection.
It introduces multi-view inconsistencies~\cite{Schwarz2020NEURIPS,Chan2020ARXIV}, but the model becomes more robust against wrong cameras~\cite{Niemeyer2021CVPR}.
In contrast, our model is able to learn the camera distribution and achieves 3D consistent results without the need of a learnable projection. 
We conclude that learning a camera generator jointly with the image generator is indeed beneficial for obtaining high-quality 3D-aware image synthesis, in particular if the camera distribution is unknown.

\boldquestion{What does the camera generator learn}
\begin{figure*}
    \centering
    \begin{subfigure}[h]{.45\linewidth}
        \includegraphics[width=\linewidth]{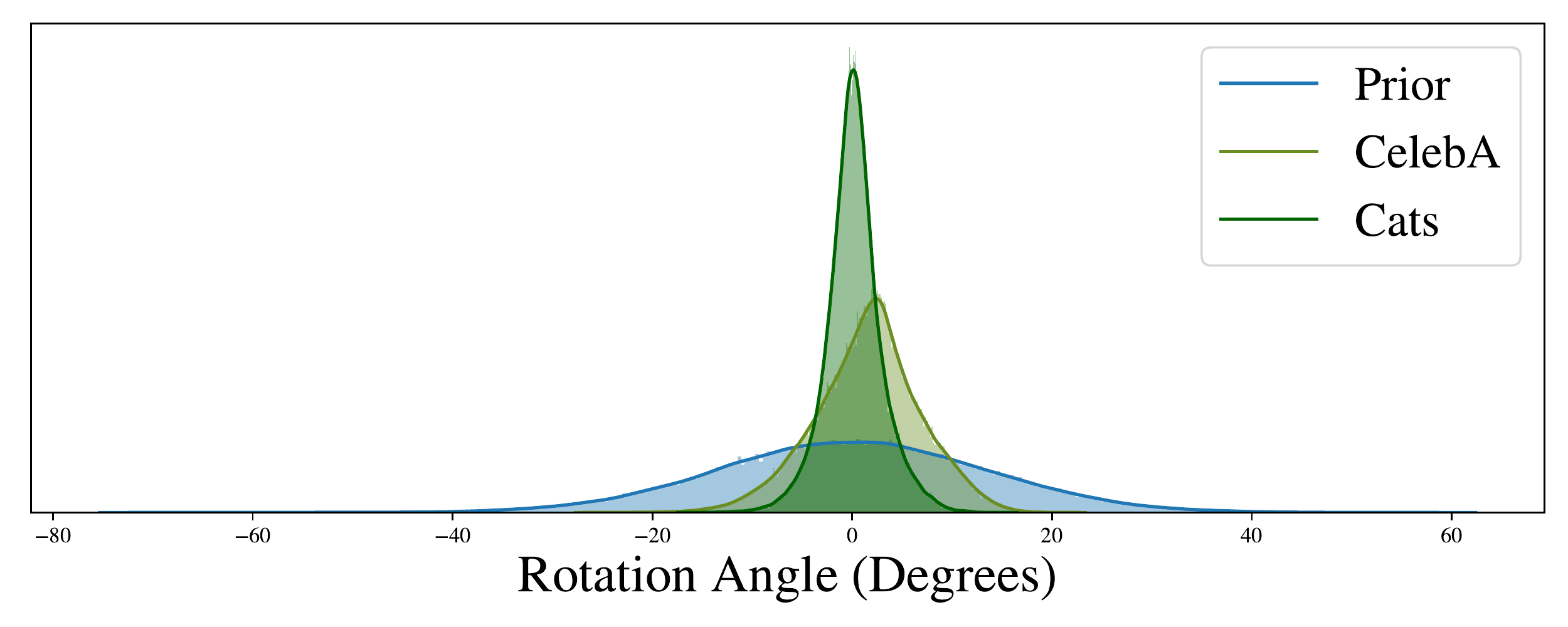}
        \vspace{-.55cm}
        \caption{Rotation Distributions for CelebA and Cats}
    \end{subfigure}
    \hfill
    \begin{subfigure}[h]{.45\linewidth}
        \includegraphics[width=\linewidth]{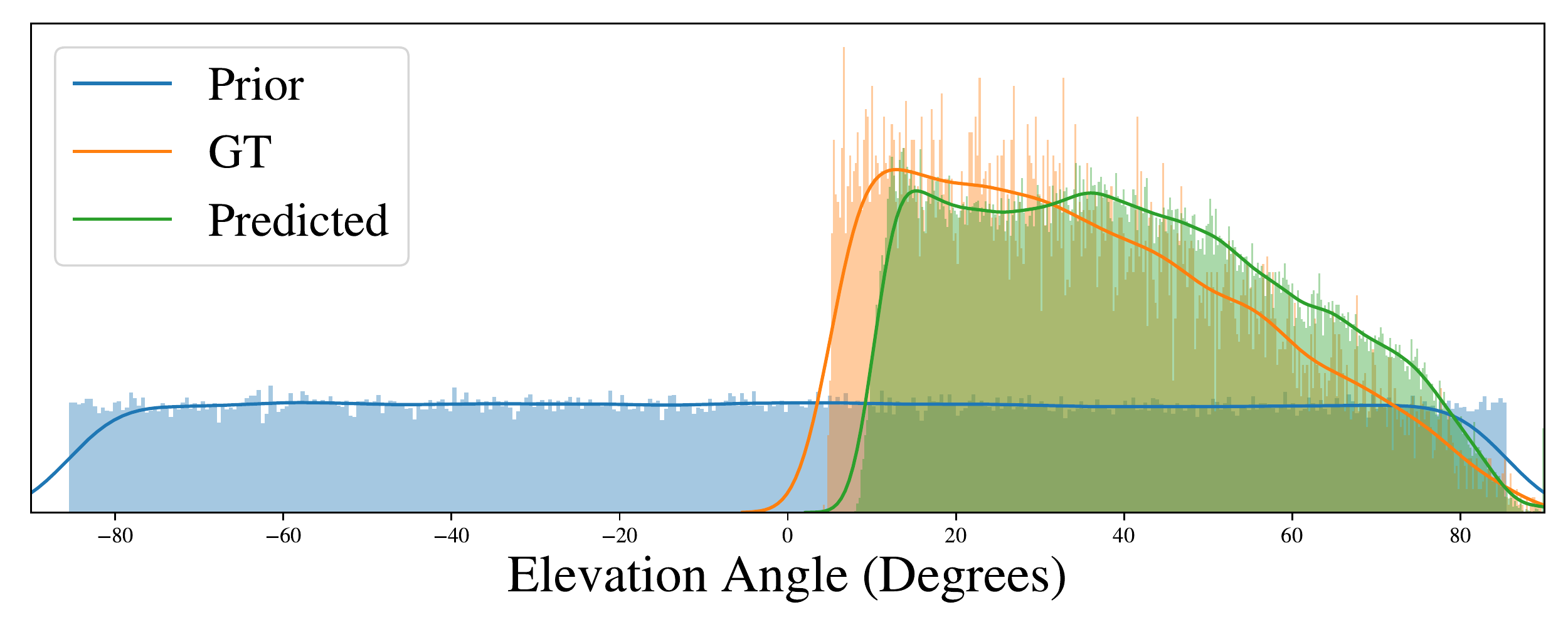}
        \vspace{-.55cm}
        \caption{Elevation Distributions for Carla}
    \end{subfigure}
    \begin{subfigure}[h]{\linewidth}
        \includegraphics[width=.45\linewidth]{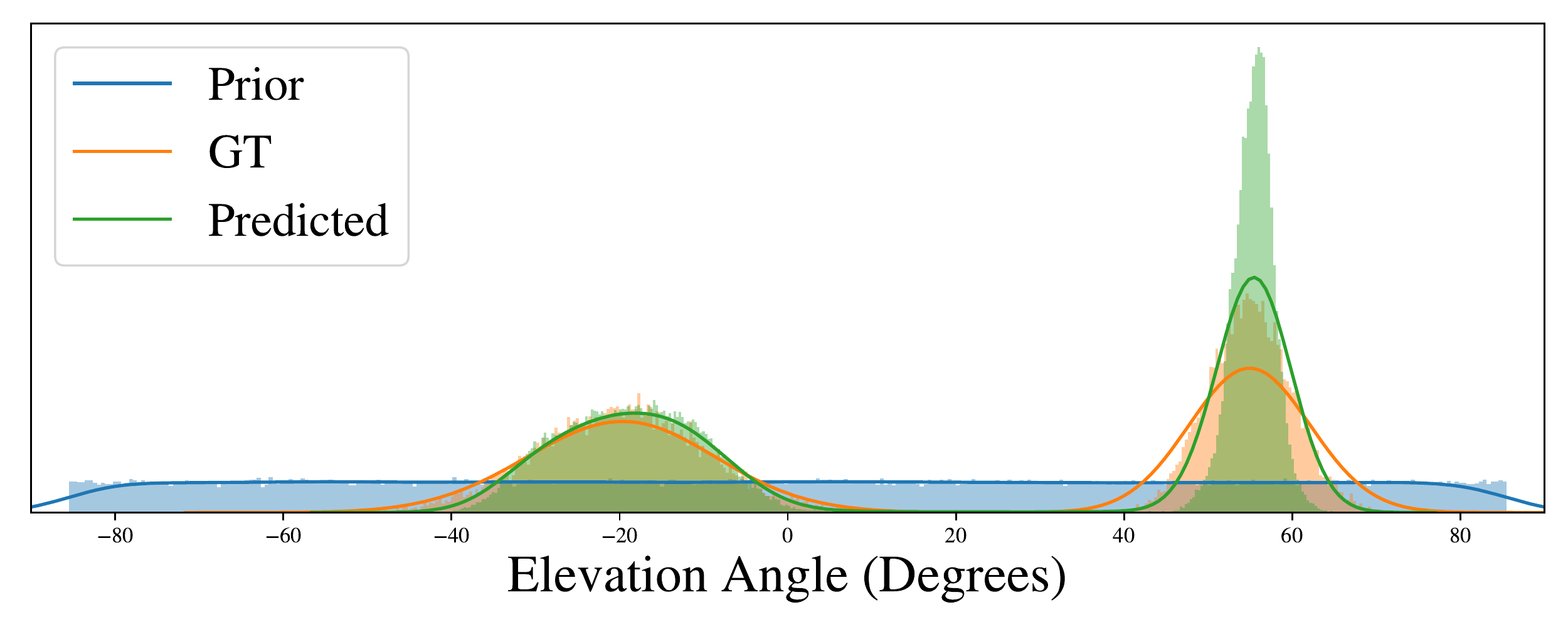}
        \hfill
        \includegraphics[width=.45\linewidth]{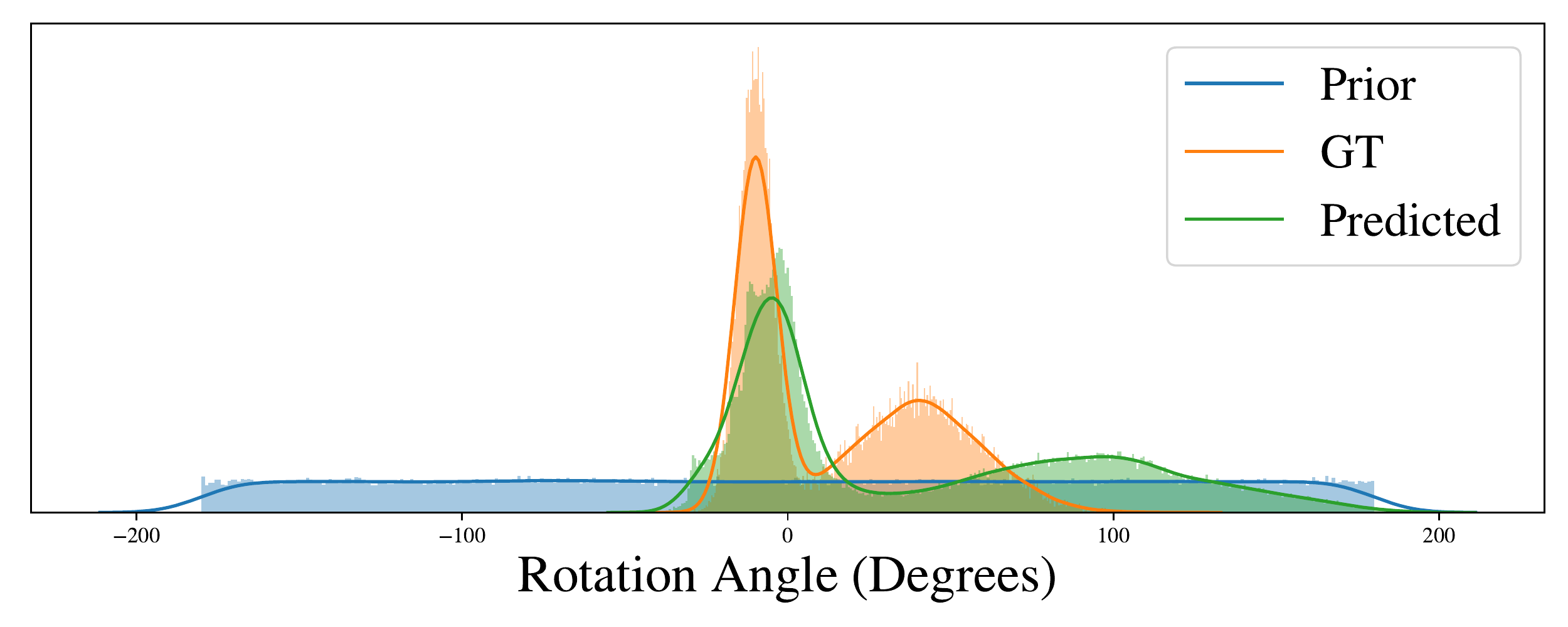}
        \vspace{-.2cm}
        \caption{Rotation and Elevation Distributions for Chairs1}
    \end{subfigure}
    \begin{subfigure}[h]{\linewidth}
        \includegraphics[width=.45\linewidth]{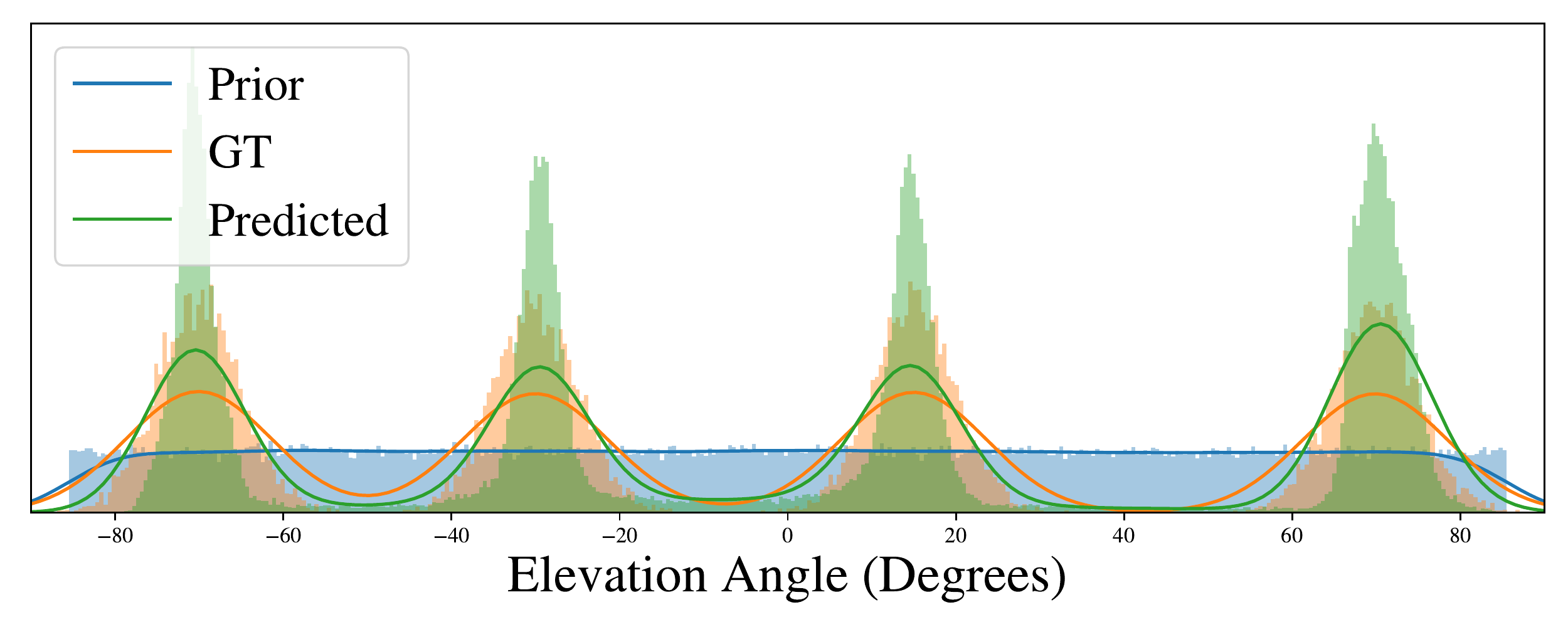}
        \hfill
        \includegraphics[width=.45\linewidth]{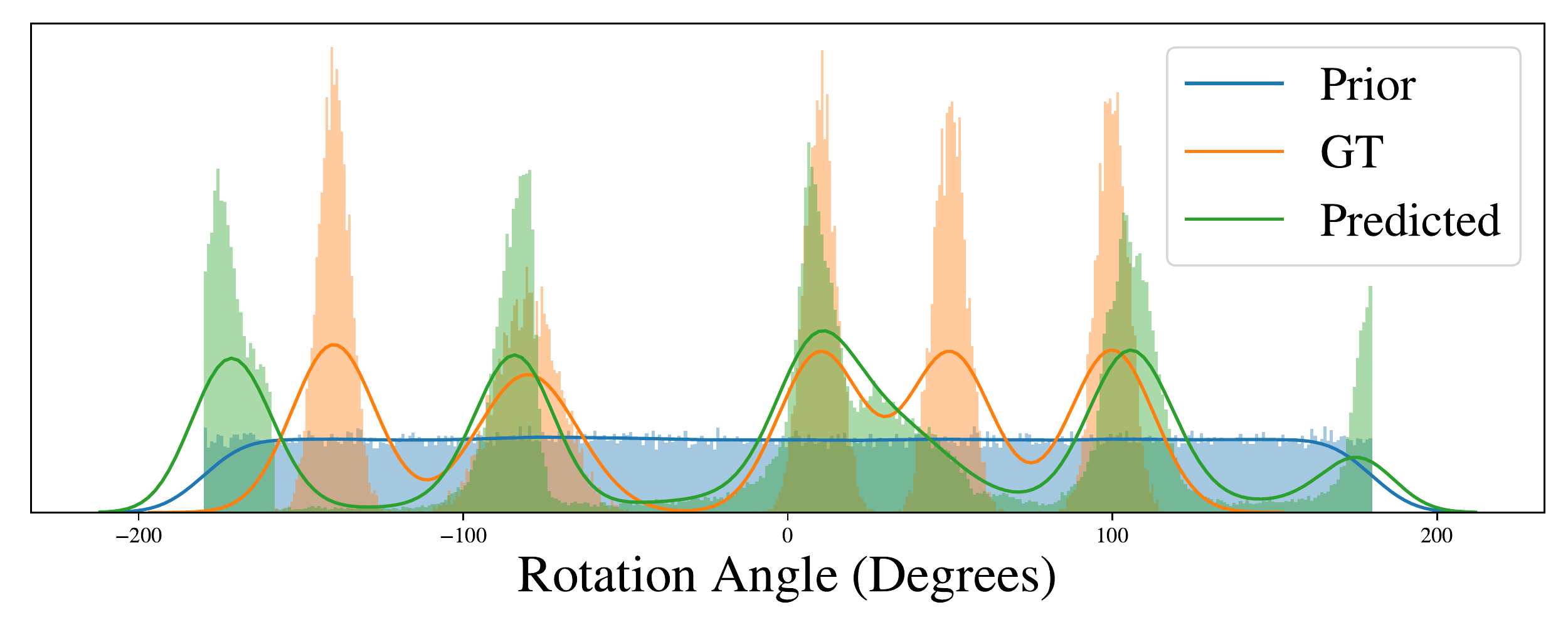}
        \vspace{-.2cm}
        \caption{Rotation and Elevation Distributions for Chairs2}
    \end{subfigure}
    \vspace{-.1cm}
    \caption{
        \textbf{Learned Cameras.}
        We show the prior, ground truth (if existing), and predicted camera elevation and rotation distributions.
        Although only training from raw image collections without any annotation, our camera generator learns to deform the prior to match the data distributions.
        Note that as the camera distributions are learned fully unsupervised, arbitrarily-shifted rotation distributions are equally-valid solutions, and here we therefore manually aligned them for this visualization.
    }
    \label{fig:camera-pose}
\end{figure*}
In~\figref{fig:camera-pose} we visualize learned camera distributions.
We can see that our camera generator learns to deform the prior distribution in a meaningful way.
For example, while starting with a uniform prior over the entire rotation and elevation range, our model adapts the camera elevation to lie on the upper hemisphere for the Car dataset as it only contains positive camera elevation angles.
Further, our model correctly approximates the more complicated marginals of the Chairs datasets.
We observe that for elevation, the predicted distribution is closer to the ground truth than for rotation.
First, arbitrarily shifted rotation distributions are equally-valid solutions as we learn them unsupervised.
Further, we hypothesize that object symmetry causes structural changes to be more dominant along the upward direction, and as a result, the model is enforced more strongly to match the correct elevation distribution.
It is important to note that as we train our model from raw, unstructured data collections, inferring the correct camera distribution is performed completely unsupervised.

\boldquestion{How important is the camera generator}
\begin{table}
    \centering
    
\begin{tabular}{ccccc}
    \toprule
    Ours & -Cam.\ Gen.\ & - BG & +Patch Dis.\ \\
    \midrule
    28 & 54 & 41 & 40\\
    \bottomrule
\end{tabular}

    \caption{
        \textbf{Ablation Study.}
        We compare FID ($\downarrow$) for our full model to ours without a camera generator (-Cam.\ Gen.) and without a background model (-BG) on CelebA at $128^2$ pixels.
        Further, we report results for training our model using a patch discriminator~\cite{Schwarz2020NEURIPS} instead of progressive growing.
    }
    \label{tab:ablation}
\end{table} 
In~\tabref{tab:ablation} we report results for our method with and without a camera generator.
We observe that learning a camera generator jointly with the image generator leads to improved results.
Qualitatively, we find that the results of our full model are more 3D consistent compared to using a camera generator without the residual design.
The residual design encourages exploration of wider pose ranges and avoids local minima of small rotation and elevation ranges (see~\figref{fig:residual_cam}).

\boldquestion{Does our model disentangle factors of variation}
\begin{figure*}
    \begin{subfigure}[h]{\linewidth}
        \includegraphics[width=.49\linewidth]{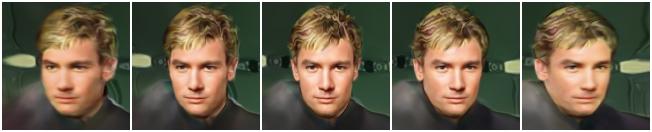}\hfill
        \includegraphics[width=.49\linewidth]{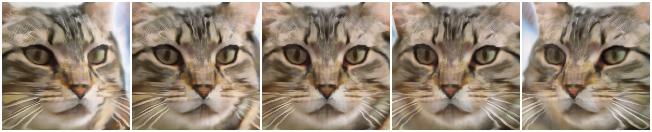}
        \includegraphics[width=.49\linewidth, trim=0 20 0 0, clip]{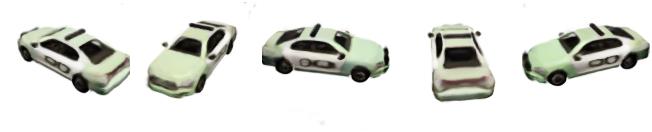}\hfill
        \includegraphics[width=.49\linewidth]{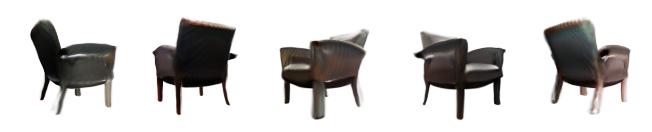}
        \subcaption{Camera Rotation}\label{subfig:cisa}
    \end{subfigure}
    \begin{subfigure}[h]{\linewidth}
        \includegraphics[width=.3\linewidth]{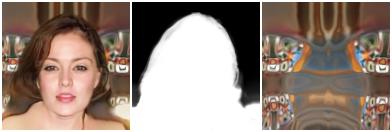}\hfill
        \includegraphics[width=.3\linewidth]{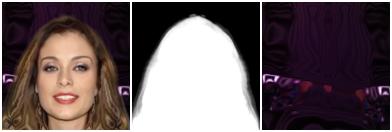}\hfill
        \includegraphics[width=.3\linewidth]{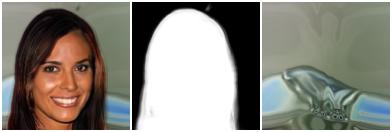}
        \caption{Fore- and Background Disentanglement}\label{subfig:cisb}
    \end{subfigure}
    \begin{subfigure}[h]{\linewidth}
        \includegraphics[width=.49\linewidth]{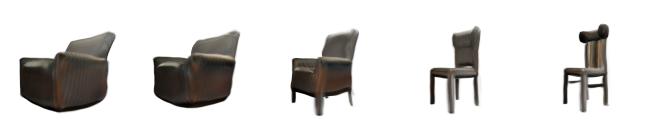}
        \hfill
        \includegraphics[width=.49\linewidth]{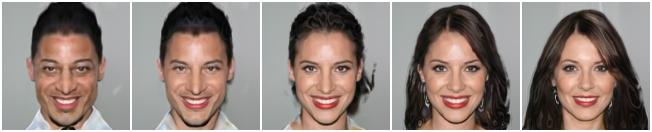}
        \caption{Shape Interpolation}\label{subfig:cisc}
    \end{subfigure}
    \begin{subfigure}[h]{\linewidth}
        \includegraphics[width=.49\linewidth]{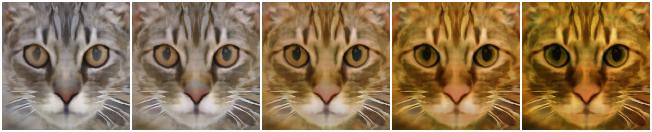}\hfill
        \includegraphics[width=.49\linewidth, trim=0 20 0 0, clip]{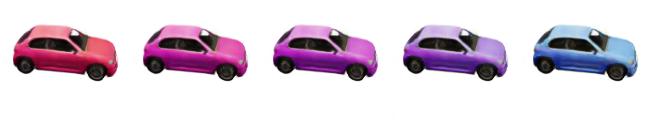}
        \caption{Appearance Interpolation}\label{subfig:cisd}
    \end{subfigure}
    \vspace{-.3cm}
    \caption{
        \textbf{Controllable Image Synthesis.}
        While training from raw image collections, we learn a 3D- and camera-aware generative model which allows for controllable image synthesis.
        We can control the camera viewpoint (\ref{subfig:cisa}), disentangle fore- and background (\ref{subfig:cisb}), and manipulate the shape (\ref{subfig:cisc}) and appearance (\ref{subfig:cisd}) of the object (see sup.\ mat.\ for more examples).
    }
    \label{fig:cis}
\end{figure*}
In~\figref{fig:cis} we show examples for how our model disentangles camera viewpoint, fore- from background as well as the shape and appearance of the foreground object.
We observe that our incorporated scene representation indeed leads to disentangled representations of different factors of variation.
At test time, these factors can explicitly be controlled, facilitating controllable image synthesis.
Although not being the primary goal, we further find that using separate fore- and background models also improves results quantitatively (see~\tabref{tab:ablation}).

\boldquestion{How important is our training regime}
In~\tabref{tab:ablation}, we compare our progressive growing regime against the patch-based training from~\cite{Schwarz2020NEURIPS}.
We find that our training regime leads to better quantitative results.
Qualitatively, we observe that fore- and background disentanglement is less consistent and the camera generator less stable for patch-based training due the reduced receptive field. %

\subsection{Limitations and Future Work}
\boldparagraphnovspace{Image Quality vs.\ Camera Exploration}
We tackle the problem of learning a generative model of 3D representations solely from raw, unposed image collection using an adversarial loss.
Note that the discriminator loss is based on 2D renderings of our model, and hence 3D consistency is purely a result of the incorporated bias.
We find that at later stages of training on CelebA, our model tends to reduce the camera pose range in favor of only increasing image quality.
We ascribe this to the large data complexity and resulting training dynamics at higher resolutions.
In practice, we avoid this tradeoff by keeping the camera generator fixed for later stages of training on CelebA.
We identify exploring how the model can be encouraged to explore the largest possible camera ranges as promising future work.

\boldparagraph{Multi-View vs.\ 3D Consistency}
\begin{figure}
    \begin{subfigure}[h]{\linewidth}
        \includegraphics[width=\linewidth]{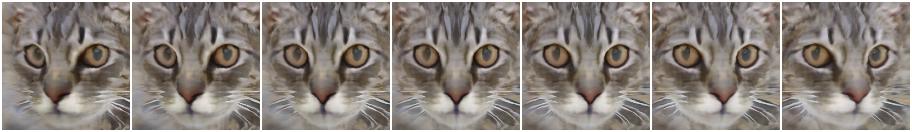}
        \includegraphics[width=\linewidth]{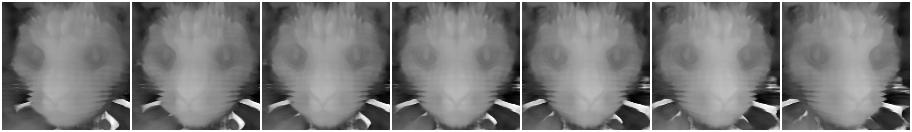}
        \subcaption{RGB (top) and Depth (bottom) for Forward-Facing Face}\label{subfig:failurea}
    \end{subfigure}
    \begin{subfigure}[h]{\linewidth}
        \includegraphics[width=\linewidth]{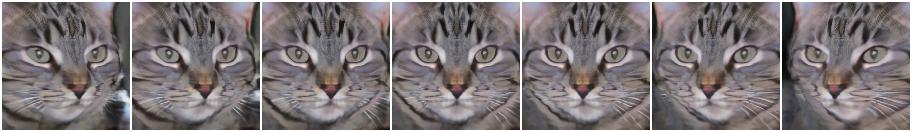}
        \includegraphics[width=\linewidth]{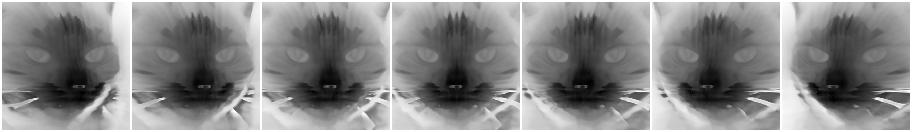}
        \subcaption{RGB (top) and Depth (bottom) for Inward-Facing Face}\label{subfig:failureb}
    \end{subfigure}
    \vspace{-.3cm}
    \caption{
        \textbf{Hollow Face Illusion.}
        Next to the expected behavior (\ref{subfig:failurea}), our model sometimes generates inward-facing faces (\ref{subfig:failureb}) (``hollow face illusion''). 
        As we train from raw image collections, this solution is equally-valid as it leads to similar multi-view consistency for face datasets.
    }
    \label{fig:failure-face}
\end{figure}
Similar to~\cite{Schwarz2020NEURIPS,Chan2020ARXIV}, we find that our model sometimes generates 3D representations which are multi-view consistent from the learned pose ranges, but not as expected, \eg, we observe ``inverted faces'' which is also known as hollow face illusion (see~\figref{fig:failure-face}).
We plan to investigate how stronger 3D shape biases can be incorporated into the generator model.

\section{Conclusion}

In this work we present \textit{CAMPARI}, a novel method for 3D- and camera-aware image synthesis.
Our key idea is to learn a camera generator jointly with a 3D-aware image generator.
Further, we decompose the scene into fore- and background leading to more efficient scene representations.
While training from raw, unstructured image collections, our method faithfully recovers the camera distribution and at test time, we can generate novel scenes with explicit control over the camera viewpoint as well as the shape and appearance of the scene.

\section*{Acknowledgment}
This work was supported by an NVIDIA research gift.
We thank the International Max Planck Research School for Intelligent Systems (IMPRS-IS) for supporting MN.
AG was supported by the ERC Starting Grant LEGO-3D (850533) and DFG EXC number 2064/1 - project number 390727645.

{\small
\bibliographystyle{ieee_fullname}
\bibliography{bibliography_long,bibliography,bibliography_custom}
}

\end{document}